# Unified Optimization of Traffic Signal and Lane Channelization for an Isolated Intersection Using Reinforcement Learning

**Jiwon Kang**
Department of Urban Engineering
Chung–Ang University, 84 Heukseok-ro, Dongjak-gu, Seoul, Republic of Korea, 06911
Email: jane3817@cau.ac.kr

**Jiyoung Kim**
Department of Smart City
Chung–Ang University, 84 Heukseok-ro, Dongjak-gu, Seoul, Republic of Korea, 06911
Email: rlawldud08@cau.ac.kr

**Hyunseok Jun**
Department of Smart City
Chung–Ang University, 84 Heukseok-ro, Dongjak-gu, Seoul, Republic of Korea, 06911
Email: jish7518@cau.ac.kr

**Keemin Sohn***
Department of Smart City, Department of Urban Engineering
Chung–Ang University, 84 Heukseok-ro, Dongjak-gu, Seoul, Republic of Korea, 06911
Email: kmsohn@cau.ac.kr

**Corresponding Author*

## ABSTRACT

Optimizing traffic signal control has been a long-lasting critical issue in traffic engineering, and a number of robust methodologies have emerged to solve the problem so far. On the other hand, the road design for lane channelization is a static problem that is difficult to solve jointly with the real-time traffic signal control. If traffic volume does not vary, it is possible that both problems can be solved in an integrated manner as a single static optimization problem. Recently, researchers have devised a mixed integer linear problem (MILP) to jointly determine the traffic signal and road width designs. However, this joint optimization problem must be solved repeatedly whenever the traffic flow pattern varies. Solving the MILP may also take impractical computing time as the problem size gets larger. We present a robust methodology for solving the joint optimization problem using reinforcement learning (RL) on a real-time basis. Abrupt changes in lane markings with an incompatible signal phase can jeopardize drivers. To sidestep such a dangerous action change, we train a model with human feedback to filter infeasible action transitions in a rule-based manner. We devised a deep Q-network that can accommodate joint actions for traffic signal and lane marking plans. To ensure the implementation of the proposed operation, a variable sign for lane markings is necessary at each stop line of an intersection. The experimental results confirm that the proposed methodology enhances the intersection performance by increasing throughput.

## INTRODUCTION

The traffic signal control problem (TSCP) has been treated independently of the road design. The road design is regarded as a prerequisite for TSCP, since researchers thought that road configuration cannot be changed on a real-time basis. The intersection capacity, however, would enhance considerably if the traffic signal control were incorporated with the dynamic lane channelization. In principle, for a fixed road width, the lane channelization can be jointly optimized along with traffic signal timings. The relevant studies have begun to emerge in the transportation research field assuming the connected and autonomous vehicle (CAV) technology is fully available (*1-3*), although any joint optimization example of traffic signal and lane channelization design cannot be seen in the real-world traffic operation. A mixed integer problem (MILP) was established for the joint optimization (*1*). This static formulation, however, is inappropriate for the conditions where traffic volumes vary often. It should be solved every time traffic conditions vary, and there is another handicap that a threshold for when to resolve the problem is not clear in the field. A rule-based method may be inadequate for identifying dynamically varying traffic conditions. Moreover, MILP has the computational burden for on-line applications, because deriving its solution depends on partial enumeration algorithms such as branch-and-bound and branch-and-price, which takes an exorbitant computing time as the problem size gets larger.

The present study suggests a robust method for joint optimization that can overcome the complications. Reinforcement learning (RL) has succeeded in engineering fields to optimize many dynamic problems with behavioral complexities. Researchers have also tried to resolve TSCPs based on diverse RL models (*4-8*). RL is the best fit to solve the proposed joint optimization problem, since traffic conditions are complex and so dynamic that a static optimization model cannot deal with it. We present a model-free RL model for the joint optimization of traffic signal and lane channelization.

Lane channelization is closely associated with traffic signal timings, but conventionally, the former is assumed to be given in advance to optimize the latter. The lane channelization with lane markings is determined at the stage of road design based on average traffic states. Using given traffic flows for each turning movement, the lane usage and markings can be determined in a static manner. Other traffic flows than the design flow do not algin with the fixed lane markings. The present study jointly optimizes the lane usage with traffic signals in an integrated framework to accommodate fluctuating traffic volumes online. Assuming the road width and number of lanes are fixed, the present study focuses only on optimizing variable lane markings along with traffic signals. The previous study optimized more attributes such as the lane width, type, and markings according to varying proportions of heavy vehicles, since it was implemented offline. The present model has an advantage to reduce the complexity in generating joint actions of lane markings and traffic signals.

We adopt a model-free RL model since the traffic dynamics cannot be simply modeled in a theoretical or rule-based way. The proposed RL model autonomously learns the traffic flow dynamics from exploration and exploitation without explicit instructions. A commercial traffic simulator, Vissim 5.0, is regarded as a virtual environment of the RL setting, and the simulation is implemented using the application programming interface (API) that Vissim provides. To be compatible with the discrete actions for lane markings, actions for traffic signals should be discrete. A deep Q-network (DQN) is set up to accommodate this discrete action set. An isolated real intersection is selected for the testbed, which resides in the Seoul metropolitan area.

The safe implementation of the proposed RL is ensured using human feedback. The dangerous switch from one joint action to another is filtered by human interception while training. Whereas human feedback facilitates the training and reduces computation time for training, we also confirm that RL without human feedback also guarantees the convergence of training by penalizing the disadvantageous action transitions that cause traffic jams with its own reward system.

The present RL model builds on the following assumptions. First, electronic signs to indicate lane markings are necessary, which should be installed on the lane surface and simultaneously mounted like a variable message sign (VMS) to reinforce the visibility of drivers. However, vehicle-to-infrastructure (V2I) and vehicle-to-vehicle (V2V) technologies could relax these strict constraints. Each vehicle can get a message regarding the lane usage from the intersection controller via communication. Second, the state of RL is defined as the queue length and the average delay in each lane of intersection. A computer vision-based surveillance system is required to collect the RL state on a real-time basis. A more rigorous assumption is necessary to compute the delay that constitutes the RL state. The delay is measured for each vehicle and aggregated on a lane basis. The conventional surveillance system is incapable of doing so, but advanced surveillance system based on artificial intelligence can track each vehicle on the road and compute the cumulative delay (*9*). V2I and V2V technologies also make it possible to trace vehicle delays (*8*).

The present paper is organized as follows. The next section reviews literature relevant to traffic signal control, lane channelization design, and their integrated optimization. The third section defines the problem and describes the testbed of an isolated intersection located in the Seoul metropolitan area. The fourth section presents experimental design and suggests metric for evaluating model performance. The modelling framework is described in the fifth section. This section describes how to set up the Markov decision process (MDP) and the corresponding RL model as well as defining state, action, reward, and discount rate. The solution algorithm is also presented in this section. The sixth section describes three reference models for comparison and scenarios for diverse traffic conditions. The seventh section provides experimental results and discusses them. The comparisons among scenarios are also presented in this section. The last section draws conclusions and suggests further extensions that could enhance the proposed model's capability.

## LITERATURE REVIEW

Previous studies are categorized in three-folds. First, traffic signal control has been intensively dealt with in traffic engineering. The history of mainstream studies is shortly introduced in this chapter. Second, the road design for lane channelization has been conducted as a static optimization problem. These studies are introduced in chronological order. Last, recent attempts to incorporate both optimization problems into a single framework are presented.

### Signal timing optimization

The TSCP seeks to determine optimal traffic signals and timings at intersections to improve traffic performance measures such as delay, throughput, queue length, and emissions. Over decades, TSCP research has evolved from fixed-time control at isolated intersections to adaptive, multi-objective control in complex networks, particularly with the advent of connected and autonomous vehicles (CAVs) and real-time data availability. TSCP literature is classified by five key dimensions (*10*): (1) network topology (isolated intersections, arterial roads, general networks), (2) control strategies (fixed-time, actuated, adaptive), (3) signal timing constraints (cycle length, phase duration/sequence), (4) road user types (vehicles, pedestrians, buses, emergency vehicles), and (5) objectives (delay reduction, throughput, fairness, environmental goals) .

Isolated intersections were tackled using classical optimization, e.g., delay is minimized under fixed cycles (*11*). Traffic signals along arterial networks were optimized with progression, and bandwidth was maximized in the MAXBAND model (*12*). General networks are addressed in later (*13*), which formulated dynamic traffic assignment and control jointly. Regarding solution methods, First, traditional models rely on mathematical programming. Mixed-integer linear programs (MILP) and nonlinear programs were applied for phase timing optimization (*14*). Second, dynamic programming (DP) emerged for real-time control, especially in small-scale networks (*15*). Last, Genetic algorithms (GAs) were applied to nonlinear, multi-objective problems (*16*).

Traffic signal control studies increasingly apply multi-agent systems (MAS), reinforcement learning (RL), and simulation-based optimization. MAS is employed by using a tabulated Q-learning based on multi-agent technology for decentralized adaptive control (*17*). Adaptive policies using traffic state feedback were devised for RL (*18,19*). Simulation-based optimization is paired with tools like VISSIM or SUMO, these allow evaluation under realistic traffic.

Regarding road users and real-time integration, TSCP models have evolved to account for heterogeneous users and constraints. Priority for transit signal priority was first modeled using actuated control (*20*). Emergency vehicle preemption and pedestrian safety were addressed in models (*21, 22*). Real-time adaptive control using sensors (inductive loops, Bluetooth, video) was a breakthrough enabling feedback policies (*23*).

Recent advances and future directions include the integration of CAV technology for signal optimization (*8*), the multi-objective optimization involving emissions and fuel (*24,25*), the meta-RL model to differentiate the objective according to varying traffic conditions (*7*), and the scalable RL methods for large networks using deep RL (*4,5*). Challenges persist in bridging theoretical models with real-world applications due to scalability, reliability of detection, and interoperability of infrastructure systems. These challenges are expected to be resolved owing to the drastic development in artificial intelligence with increasing computing capabilities.

### Lane layout design

Lane layout design has traditionally focused on optimizing lane markings or fixed lane allocations based on uniform lane widths. An early study suggested the structured three-step approach for optimizing lane allocation (*26*). Subsequent studies explored the complex interdependence between approach and departure lane configurations, particularly in relation to operational coordination and geometric continuity (*27-29*). To account for time-varying traffic conditions, a robust optimization method was also developed that incorporates demand fluctuations into lane allocation decisions (*30*). Most of these studies assume the number of lanes on each intersection leg as an exogenous input, based on standard lane width. The present study also assumes constant road and lane widths, allowing only lane markings to vary according to traffic conditions. However, the number of feasible lanes is inherently tied to the available roadway width, which is typically fixed and cannot be easily expanded in a largely fluctuated traffic condition (*31*). Lane width selection is influenced by a variety of factors, including designated design speeds, vehicle dimensions, and lateral operating characteristics such as vehicle sway and oscillation envelopes (*32-34*).

In practice, the minimum lane width is commonly established based on design speed, following national or local geometric design standards. Yet, in dense urban environments with closely spaced intersections and chronic congestion, prevailing speeds are often significantly lower than the design speed, opening possibilities for narrower lanes without compromising safety (*35*). Conventional standards prioritize safe accommodation of the largest design vehicles—typically heavy trucks and buses—which often results in excessive lane widths for the dominant vehicle class, namely passenger cars. Empirical studies show that lateral occupancy on multilane roadways frequently hovers just above 50%, reflecting underutilization of lane space (*36*). Moreover, heavy vehicle share in urban traffic is relatively small, further contributing to lateral inefficiencies and elevated infrastructure costs (*37*). In this context, the present study fixes the lane width at a minimal level, while adjusting only lane usages on-line along with traffic signals.

**Integrated optimization**

The traffic signal control cannot be easily integrated with the lane design problem, because the former is, in nature, dynamic and on-line, whereas the latter is static and off-line. Nonetheless, a recent study presented an incorporated framework to simultaneously optimize both problems (*1*). This recent study formulated a MILP to optimize the lane layout, including lane number, type, width, and allocation, along with traffic signal timings. These control variables were simultaneously optimized to enhance intersection capacity. The problem adopts constant traffic flows as input and returns the optimal lane usages and signal timings. This preceding study inspires us to develop an online joint optimization model and is adopted as a reference to validate the model.

## PROBLEM DEFINITION

### Problem set-ups

This subsection defines the problem to be solved in a concrete way. RL actions for TSCP can be categorized into two types. The first action type handles discrete signal phases (*38-40*). At the beginning of every decision interval, the best signal phase is chosen out of the candidate signal phases. To do so, candidate signal phases that are exclusive from one another should be predefined. The decision interval must be fixed. Yet, the same phase can be selected consecutively, allowing for flexible signal timing with multiples of the interval. The second action type adopts phase durations as decision variables (*17,18,41*), wherein a sequence of traffic phases must be fixed. The cycle length is autonomously altered in terms of chosen durations. The present study adopts the former discrete action type for an isolated 4-leg intersection such that a signal phase is chosen among four predefined signal phases every 20 seconds, which facilitates to be combined with lane marking actions that are intrinsically discrete.

Actions for the lane channelization are confined to altering lane markings, while fixing the number and width of lanes. Another restrict is inflicted that each intersection leg must have at least one straight-through and left-turn lane, respectively and that all rightmost lanes allow the right-turn and straight-through traffic. Under these conditions, the number of feasible lane-marking cases is reduced to 3 for each 4-lane approach (see Fig. 1(b) below). The present study allows only changes between the three cases in lane markings for an approach, such that drastic changes are excluded. These lane marking actions should be combined with traffic signal actions, and the total number of combined actions is 36, details of which will be described in the next section.

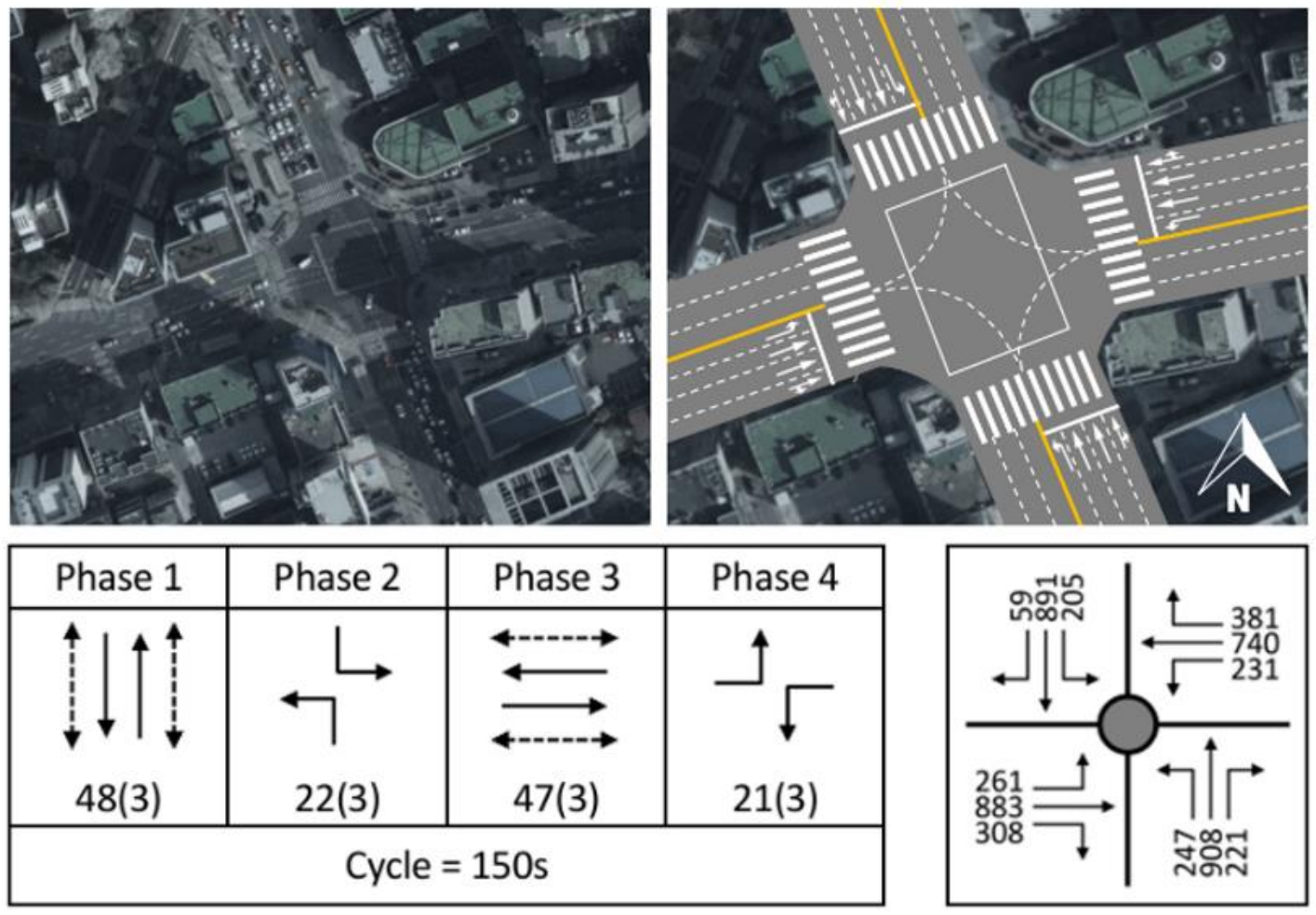

(a) Testbed of a real intersection in Seoul

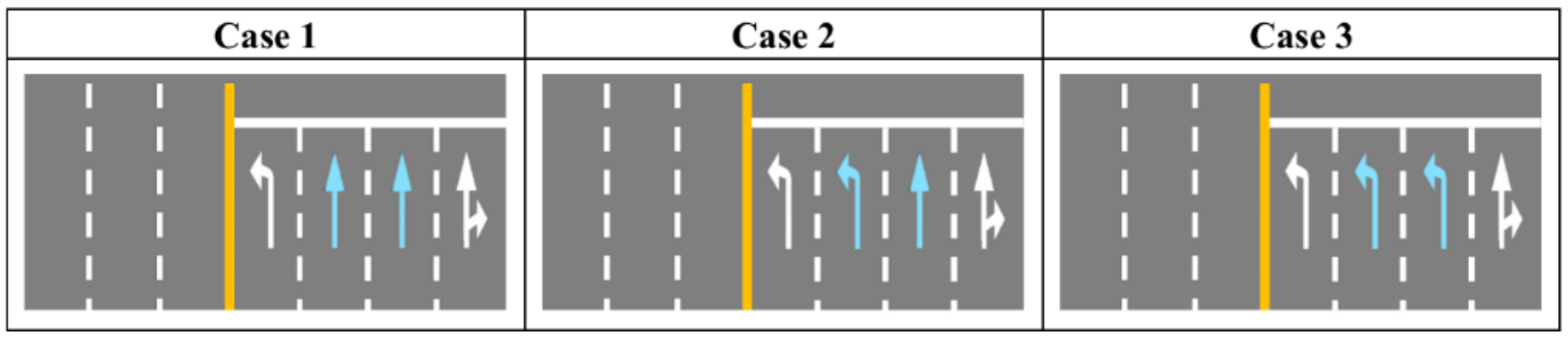

(b) Three allowable lane marking cases for an approach of the testbed intersection

**Figure 1** Testbed and experimental environment

An additional treatment is necessary to maintain the feasibility of combined actions. Even though each combined action is feasible, sequentially implementing arbitrarily chosen two actions can endanger drivers. An abrupt change in lane markings and traffic signals may force drivers to escape from the current lane to an adjacent lane compatible with the new action. There are two options to handle such a situation in this study. First, the dangerous action changes are filtered by human feedback while training. This is more effective than the predefinition of action changes in danger. The second way is to trust the RL reward mechanism that can autonomously penalize dangerous implementations. Vehicles to forcefully change lanes induce a traffic jam by blocking traffic behind them, which then leads to receiving negative rewards in RL. All remaining experiments are conducted under both conditions.

**Testbed set-ups**

The testbed for validating the proposed joint RL is selected from the Seoul metropolitan area. Fig. 1 shows the 4-leg intersection with 4 lanes for each approach and its actual traffic signal phases. Actual signal phases and timings are also demonstrated in Fig. 1. Traffic volumes for each turning movement can be seen in the bottom-right corner of Fig. 1(a). Treating these traffic volumes that lead to LOS "C" as baseline, we synthesized two sets of different traffic volumes: one for a congested condition (LOS "E") and the other for a free-flow condition (LOS "A"). For each of the three traffic volume scenarios, we randomly alter every directional traffic flow by up to $\pm 20\%$ every 20 minutes during each 1-hour episode while training, which fluctuates the turning proportion of traffic.

Although the proposed methodology is tested for a single isolated intersection, it can be used in a large road network with a multi-agent RL (MARL). A MARL can adopt the proposed RL model as each agent and make them cooperate with each other, which can be conducted without infringing the basic modelling framework. Once the computing resources are secured in the future, the proposed RL model could be easily extended to an environment of area-wide road network.

## MODELLING FRAMEWORK

This chapter describes how to associate the problem established in the previous section with an MDP for RL. The definition of each element in the MDP is presented, and a DQN for approximating the true action-value function is set up. The algorithm to implement the RL model is provided.

### Markov Decision Process (MDP)

MDP is the target to be addressed via RL and is represented by a tuple of 5 elements as denoted in **Equation 1**.

$$MDP = \langle S, A, T, R, \gamma \rangle \quad (1)$$

$S$ is a set of states, $A$ is a set of actions, $T$ represents a model of transition function, $R$ is a reward function, and $\gamma$ is a discount rate for the future reward. Solving MDP via RL means that an optimal policy function is derived from experiments by maximizing the cumulative sum of discounted future rewards, which is called the return ($G$) expressed by **Equation 2.**

$$G = \sum_{t=0}^{\infty} \gamma^t R_{t+1} = R_1 + \gamma R_2 + \gamma^2 R_3 + \gamma^3 R_4 \cdots \quad (2)$$

More specifically, for Q-learning that is adopted in the present study, the optimal action-value function (Q-function) is derived from experiments, given the sets of available actions and states. The current state transitions into a next state after taking an action, and the probability of this transition constitutes the model of MDP. Unfortunately, the model is unknown in the present study, so we adopt a model-free Q-learning method that regards the transition mechanism as implicit. The mechanism, however, can be unveiled by experiences while learning.

### Definition of state, action, reward, and discount rate

The state of RL environment represents the traffic condition of the testbed. Two measures of queue length and delay define the state. These two measures are collected for each lane at every decision interval. We assume the vision-based traffic surveillance system can collect both measures. Consequently, the dimension of state is $2 \times 16$, where each of 4 incoming approaches has 4 lanes.

An action of the present MDP includes all possible pairs of traffic signal phases and lane marking options. As mentioned earlier, we adopt the discrete action set to utilize the DQN in the original form (*42*). Given four different signal phases in a single ring and three possible lane marking options for each intersection approach, the RL agency chooses a single phase with two lane-marking options, each for an approach included in the phase. We set up the decision interval at 20 seconds and insert 3-second amber time between changing phases. When the same signal phase is selected as the previous one, the amber time is omitted. The dimension of an action is $1 \times 36$, since the number of candidate signal phases is 4, the number of lane-marking options is 3 for each approach, and a signal phase accommodates 2 incoming approaches ($36 = 4 \times 3^2$).

The reward definition is the most important in RL, since it reflects the learning objective. Maximizing the return is the main target of RL. The reward should be defined such that after taking an action the RL agency can get a reward when the intersection performance ameliorates and a penalty when the intersection performance deteriorates. The intersection performance can be measured by conventional indices such as delay and throughput. We select the throughput as the baseline performance measure.

The present study adopts the discrete reward specification established in the original DQN (*42*). That is, the reward is set at +1 when the throughput increases and -1 when the throughput decreases, after taking an action every interval. This simple reward specification outperforms others that directly quote the performance measures.

The last element of MDP is the discount rate that weighs future rewards to derive their present values. The rate is set at 0.99, since the same value was used in previous successful implementations of RL in traffic signal controls (*4-8*).

**Deep Q-network**

We employ specifications of the original DQN to solve the proposed joint optimization problem (*42*). The action-value function (Q-function) is the target to be estimated. **Equation 3** denotes the Bellman optimality equation that is used to estimate the optimal Q-function, and correspondingly the optimal Q-function plays a role of policy function $\pi(s)$ that returns an optimal action given the current state. **Equation 4** explains how the optimal Q-function returns the best action. $s$ denotes a state vector that belongs to $S$, $a$ is an action vector that belong to $A$, and $r$ is an immediate reward given after the action $a$ is taken. The immediate reward takes $\pm 1$ as described above.

$$Q(s,a) = \mathbb{E}(r + \gamma \max_{\boldsymbol{a'}} Q(s,a')) \tag{3}$$

$$\pi(s) = \max_{\boldsymbol{a'}} Q(s,a') \tag{4}$$

The parameterized Q-network, $Q(s,a|\theta)$, which is a neural network with weights $\theta$, approximates the true optimal Q-function $Q(s,a)$. **Equation 5** denotes the loss function to be minimized while training to estimate the parameter $\theta$. DQN has a trick to deal with neural parameter $\theta$ to facilitate learning (*42*). The trick separates the target Q-function, $Q(s_t, a'_t|\bar{\theta})$, from the main Q-function, $Q(\boldsymbol{s_t}, \boldsymbol{a_t}|\theta)$. The target Q-function is then updated in a longer interval, whereas the main Q-function is updated every decision interval. Fixing this target parameter ($\bar{\theta}$) for a while renders the convergence of $\theta$ faster. The algorithm to implement the minimization of the loss function will be described in the next sub-section. Intuitively, minimizing the loss function ensures that the experimental outcome can hold the Bellman optimality condition as closely as possible.

$$\mathcal{L}(\theta) = \mathbb{E}\left[\left(r_{t+1} + \gamma \max_{\boldsymbol{a'}} Q(s_t, a'_t|\bar{\theta}) - Q(s_t, a_t|\theta)\right)^2\right] \tag{5}$$

We set up a DQN for the present study. Its architecture is determined on a trial-and-error basis, after testing as many plausible candidates as possible. Fig. 2 shows the best architecture, wherein the input dimension is compatible with the state dimension, and the output dimension with the action dimension. This architecture facilitates deriving the best action by selecting an action corresponding to an output node of DQN that returns the maximum value.

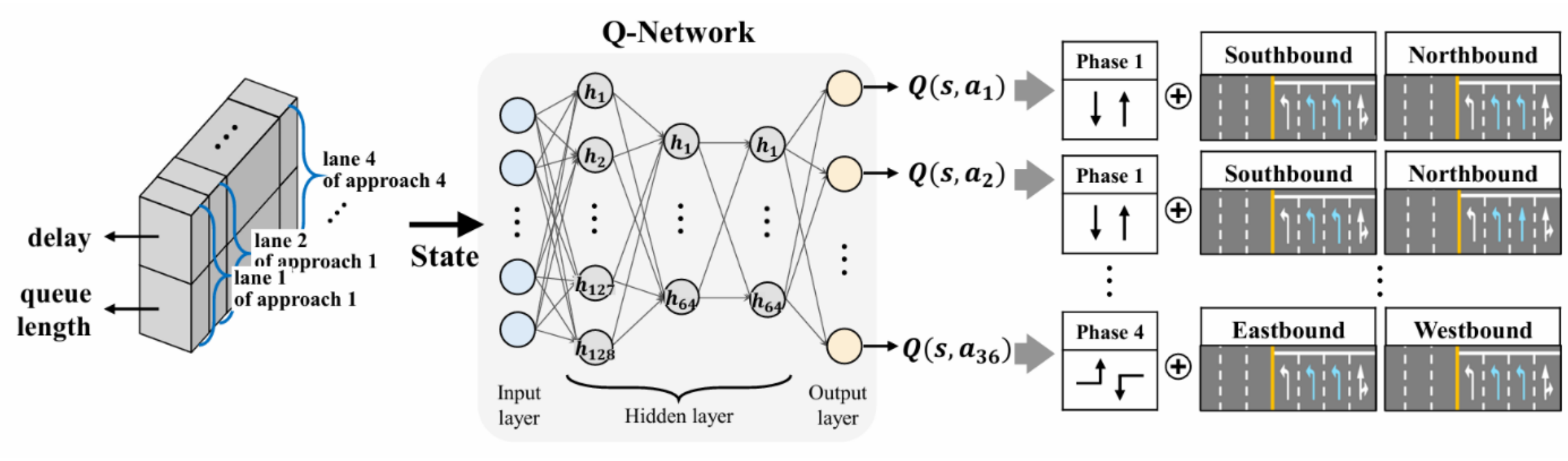

**Figure 2** DQN architecture

**Solution algorithm**

Experiments are conducted in the VISSIM 5.0 environment. The duration of each episode ($T^{max}$) is set at 1 simulation hour (=3,600s), and the number of episodes ($N$) is set at 500. The VISSIM is a time-driven simulator that updates the state of all vehicles every time interval ($\Delta$=0.1 sec). The RL decision interval ($T$) is 20 seconds. The exploration rate ($\varepsilon$) is scheduled so that it dwindles as iterations proceed. Starting at $\varepsilon_{max}$ (=100%), the rate converges to $\varepsilon_{min}$ (=10%) at the final stage of learning, by setting the constant $E$ at 800,000. The size of replay memory ($N_R$) to store transition experiences is 20,000. A transition is defined by $< s_{t-1}, a_{t-1}, r_t, s_t >$. The mini-batch size ($M$) stores transitions from experiments and is set at 64. The interval ($C$) for updating target Q-function at 40,000 seconds. The following shows the algorithm to learn the proposed MDP based on the stochastic gradient descent algorithm (SGD), in which auto-differentiation functions in TensorFlow are used for the computation of the gradient.

```
Randomly initialize θ, set θ̄ = θ
Initialize summation clock ( t_e =0)
For each episode = 1, …, N
  Initialize summation clock ( t =0)
  Set a random seed of traffic simulation
  While  t < T^max
        Update the cumulative time for exploration t_e = t_e + Δ
        Run a single step of traffic simulation  t = t + Δ
        If t + T_a  is a multiple of T then
                Update exploration rate ε
                ε = (ε_max − ε_min) e^(−(t_e/E)^2) + ε_min
        Choose a random number from the uniform distribution ranging from 0 to 1
        If the chosen random number <  ε
                Randomly select a joint action  a_t
        Else
                a_t = argmax_a Q(s_t, a|θ) .
        If the signal phase in the selected a_t differs from the previous one in a_{t−1}
                Implement amber phase for the current phase
                Repeat a single step of traffic simulation until    t = t + T_a
                Implement action  a_t
        Else
                Implement action  a_t
                Repeat a single step of traffic simulation until    t = t + T_a

        Compare the current throughput with the previous one.
         r_t=+1, if current throughput < previous throughput
         r_t=-1,  if current throughput > previous throughput
         r_t= 0,  if current throughput = previous throughput

        s_t= queue length and average delay in the current time step t for all lanes
        Store transition < s_{t−1}, a_{t−1}, r_t, s_t > into replay memory
        If replay memory size > D then
                Remove the oldest transition from replay memory
        Sample M transition experiences  < s, a, r, s' >_j randomly from replay memory
        Perform a single SGD step on
                Min. Σ_{j=1}^{M} ( r_j + γ max_{a'} Q( s'_t, a'_t|θ̄) − Q(s_j, a_j|θ) )^2  w.r.t  θ
        For every C time intervals Set  θ̄ = θ
```

## EXPERIMENTAL SET-UPS

### Reference models and scenarios

Three reference models are set up for comparison. As a baseline, the pretimed traffic signal controller that is being operated in the real world is chosen. For scenarios that alter traffic volumes, the conventional method in the highway capacity manual (HCM) determines new signal timings. Second, an RL model is implemented to control only traffic signals with lane markings unchanged. Third, the MILP model presented in the latest study (*1*) is chosen for comparison. This static model is implemented whenever the traffic flow pattern varies, ignoring the judgement of those timings. This would offer an advantage for MILP when compared to other models. Henceforth, the reference models will be called PRETIM, RLSIG, and MILP, respectively. The proposed RL method to jointly control traffic signals and lane markings is compared with these three reference models.

The traffic conditions are differentiated in terms of traffic volumes. As mentioned earlier, the observed traffic volumes are a baseline, which indicates the level of service (LOS) "C". We increase the baseline traffic volumes until the LOS reaches "E", decrease the volumes until the LOS reaches "A". For all these traffic-flow conditions, directional traffic flows are altered within $\pm$20% of the observed ones every 20 minutes during each 1-hour episode. Consequently, experiments are conducted on the different traffic-flow patterns using three different control models.

### Metrics for evaluation

Besides the average reward, a more intuitive metric is necessary to evaluate the performance and compare it with that from other models. The immediate reward is dependent on the change in traffic throughput, but the reward is not intuitively understandable. Thus, at inference time after training, the model performance is directly measured by the throughput, which can be calculated by the total number of vehicles that pass through all stop lines of intersection during unit time (=1 sec). We use this single measure, since all traffic conditions in the present experiments are under saturation. The throughput is inversely proportional to the vehicle delay until the traffic flow saturates. Comparisons among different scenarios and models are conducted based on the throughput.

## RESULTS AND DISCUSSION

### Convergence of RL

Fig. 3 demonstrates the convergence graphs that show the average reward across all episodes. For each episode, all given rewards are averaged. As episodes proceed, the average reward shows an increasing trend in moving average to a saturated value. The average reward tends to more fluctuate at the beginning of learning, and the variation stabilizes shortly as episodes proceed. The saturation level slightly exceeds 0, which is rational because our definition of reward depends on the relative enhancement in throughput with respect to the previous iteration. That is, if the saturation level approached to +1, this would mean that the throughput increases without a ceiling as iterations proceed. Such a situation is infeasible in the real world.

Regardless of exploration or exploitation, an action change may induce a situation where vehicles encounter a lane marking that is contradictory against their turning direction. There are two types of action takings in this study to avert such an irrational action change. First, irrational action changes are filtered by human feedback while training. The number of possible action changes is tantamount to $36 \times 36$ in the present experiments. Since enumerating every harmful action pair in advance is inefficient, we empirically select irrational action changes by watching animation for simulation while learning. More specifically, the action changes are stored as a transition along with states before and after action taking. The collected harmful transitions, henceforth, will be banned while learning. Such human feedback is general for RL implementations in many fields (*43-45*). Second, no human filtering is engaged in learning, so that the DQN can autonomously learn by getting a penalty for harmful switching of actions. Interestingly, the latter scheme also works well, and the gap in convergence pace between both training schemes is not that large. The convergence graph in Fig. 3 (a) reaches saturation slightly faster than that in Fig. 3 (b). We conduct the subsequent experiments based on both schemes.

When comparing the convergence over two RL models, RLSIG that accommodates only traffic signals converges slightly faster than the proposed RL model that jointly optimizes both traffic signals and lane markings. On the other hand, the computation time for inference after training shows no discrepancy between both models.

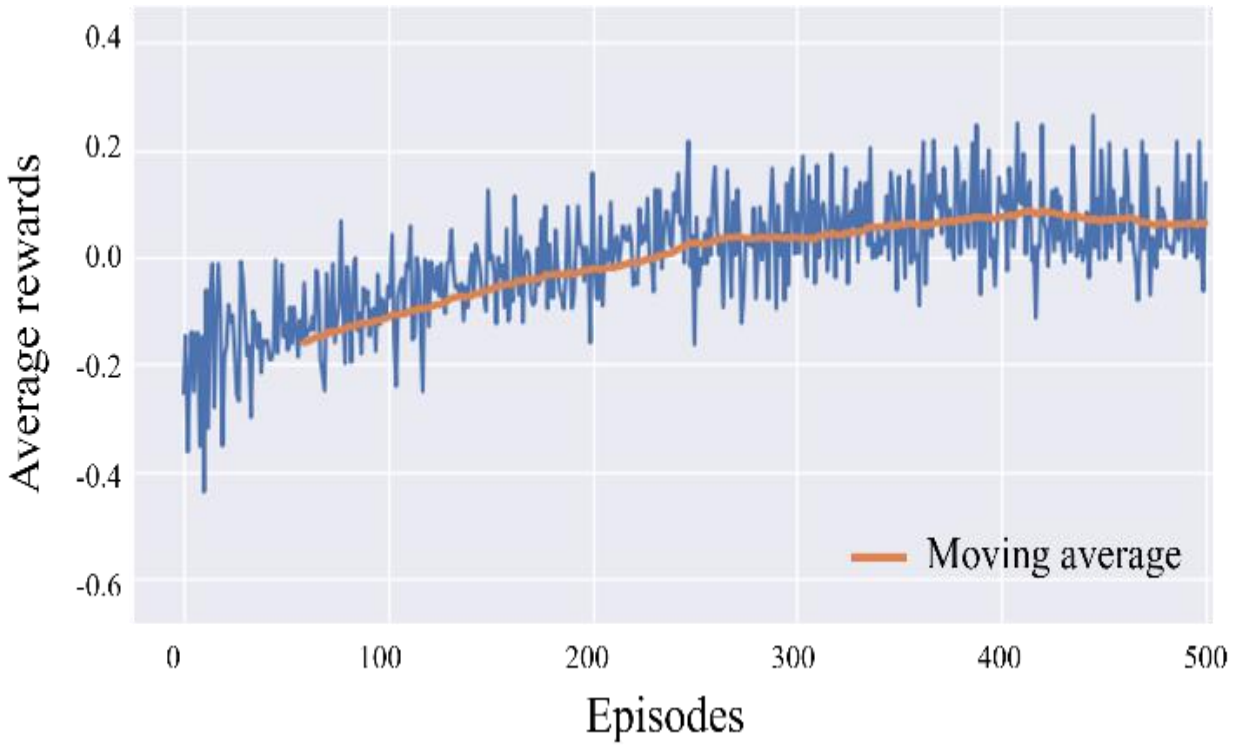

(a) The proposed RL with human action filtering

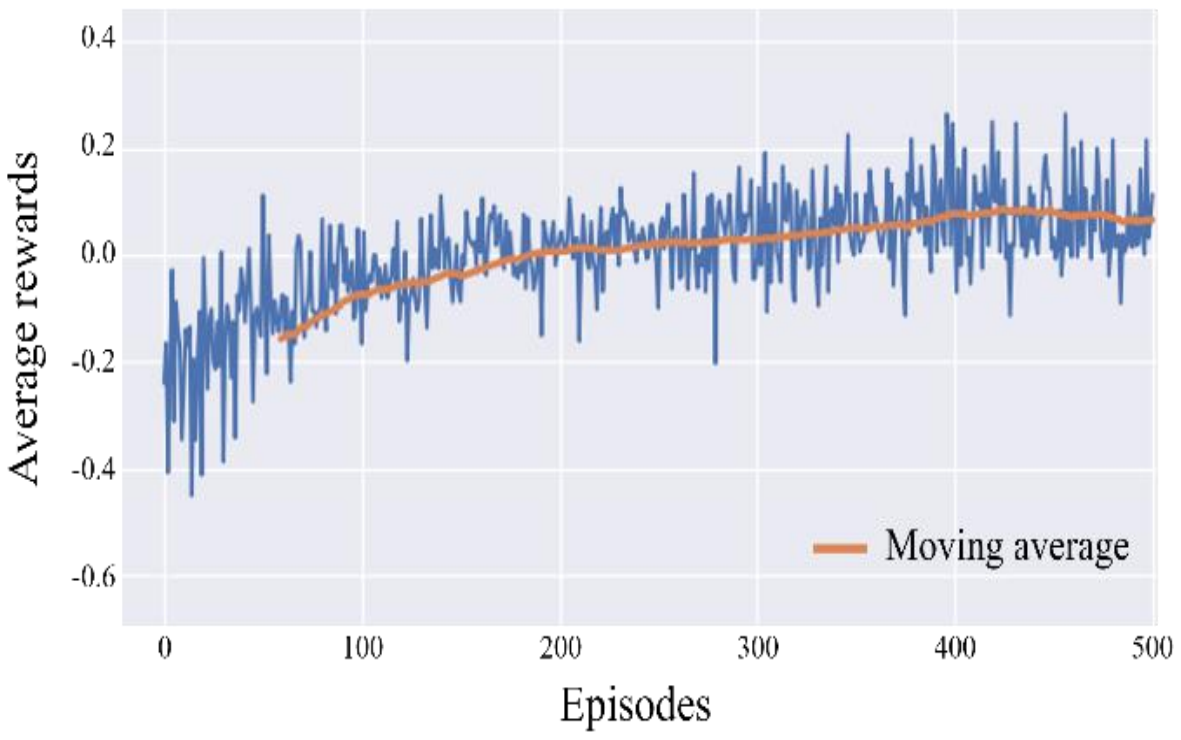

(b) The proposed RL without human action filtering

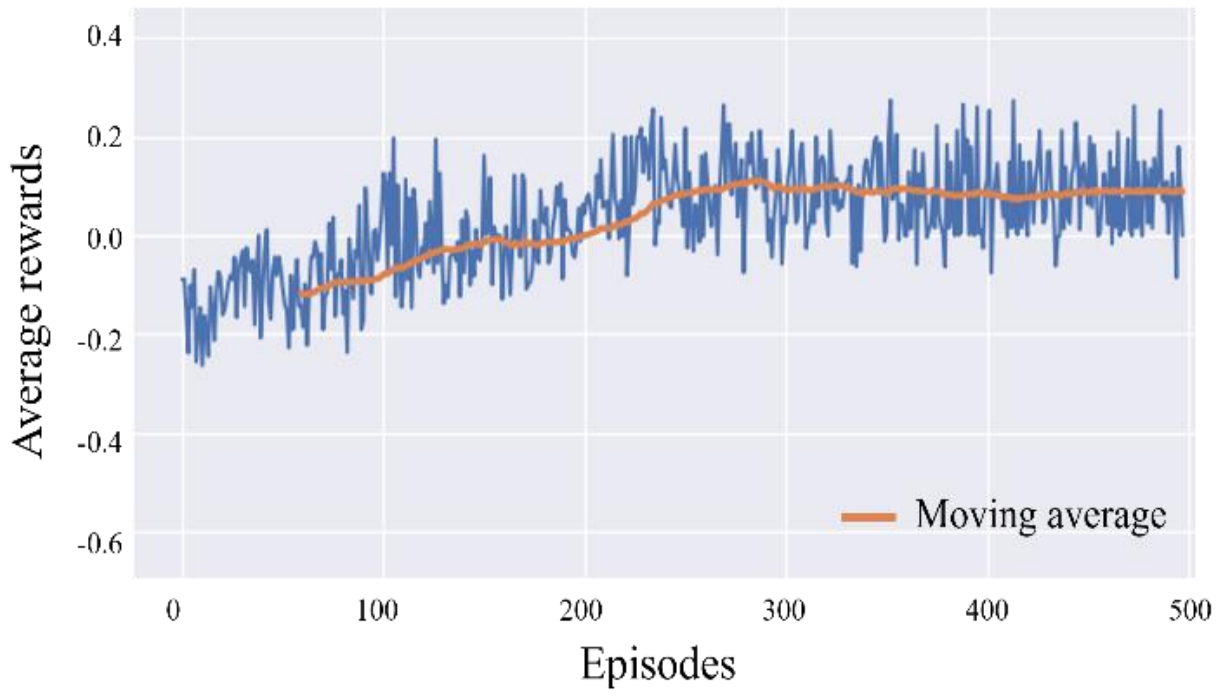

(c) The convergence of RLSIG

**Figure 3** Convergence of RL models

**Comparisons with reference models for different traffic conditions**

The performance measure of the two RL models is computed for each scenario at inference time after training. The measure of two other models is also computed separately across different scenarios. It should be noted that PRETIM and MILP cannot autonomously detect the timing of change in traffic conditions. Nonetheless, we intentionally reimplement these models whenever traffic volumes are changed in scenarios with directional traffic flows varying, we reimplement PRETIM and MILP when the traffic volumes are changed and collect the performance measure using the refreshed models. Despite this advantage of its competitors, the proposed RL demonstrates better performance by passing more vehicles in unit time than PRETIM. The performance gap gets larger as the LOS exacerbates from "A" to "E". When compared with RLSIG with only traffic signal optimization, the proposed model records a larger throughput value for each LOS. This confirms that the simultaneous optimization of traffic signals and lane markings clearly outperforms the normal RL-based traffic signal control. Regarding the comparison with MILP, both models record almost the same performance in passing vehicles. It should be noted, however, that MILP cannot showcase such an identical performance without the autonomous adjustment for changing traffic volumes. The proposed RL model shows the almost same throughput values by whether to include human feedback. Contrary to our prior expectations, the benefit of human feedback is marginal, only with a slight saving in training time. **Table 1** lists the experiment results of what are described above.

**TABLE 1 Average throughput resulted from each model across varying traffic conditions**

(vehicles/second)

| LOS (HCM-based) | Turning rate | Human Filtering | Proposed RL | PRETIM | RLSIG | MILP |
|---|---|---|---|---|---|---|
| A | Fixed | Yes | 0.74 | 0.73 | 0.73 | 0.74 |
| | | No | 0.74 | | | |
| | Varying | Yes | 0.74 | 0.72 | 0.73 | 0.74 |
| | | No | 0.74 | | | |
| C | Fixed | Yes | 1.45 | 1.41 | 1.43 | 1.46 |
| | | No | 1.45 | | | |
| | Varying | Yes | 1.44 | 1.37 | 1.41 | 1.43 |
| | | No | 1.44 | | | |
| E | Fixed | Yes | 1.43 | 1.08 | 1.20 | 1.45 |
| | | No | 1.43 | | | |
| | Varying | Yes | 1.41 | 1.06 | 1.17 | 1.41 |
| | | No | 1.40 | | | |

Fig. 4 illustrates examples to account for why the proposed RL outperforms RLSIG. If only traffic signals are considered, an abrupt surge of traffic demand in a specific turning direction, the left turn in the example, causes traffic jam even though RLSIG tries to adapt the traffic signal. Jammed vehicles block neighboring lanes due to insufficient capacity assigned for the turning direction. On the other hand, the proposed RL well adapts that situation by increasing left-turn lanes as well as optimizing the pairing traffic signal. The contrast between Fig. 4 (a) and (b) demonstrates the capability of the proposed RL model to cope with such dramatic traffic change. When RSIG is implemented, some vehicles are in the straight-through lane and try to turn left, which cause disorder by blocking other vehicles behind. On the other hand, the proposed model autonomously increases the left-turn lanes to allow those vehicles to pass when left-turn demand surges.

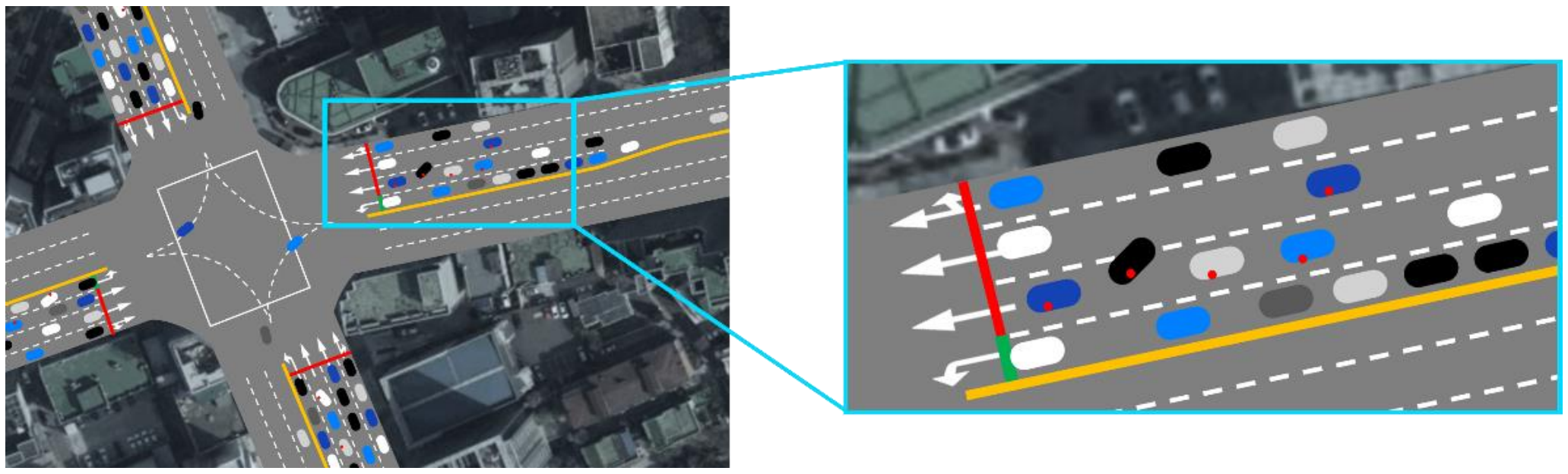

(a) RSIG – Only a single left-turn lane causes disorder due to an abrupt surge of left-turn traffic

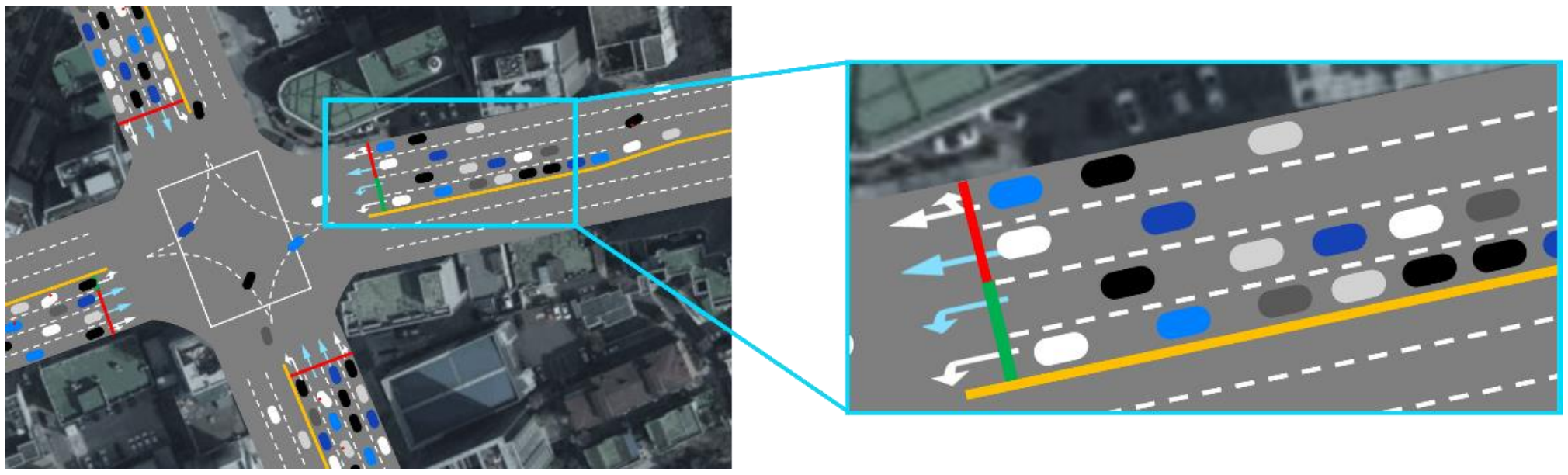

(b) The proposed RL – Two left-turn lanes are assigned online to accommodate an abrupt surge of left-turn traffic

**Figure 4** The capability to cope with an abrupt surge in directional traffic demand

Another helpful scheme to raise the performance of the proposed RL is to apply more candidate traffic signal phase. To do so, we add four more candidate phases excerpted from the dual-ring signal system. This addition enhances model performance. The average throughput increases to 1.47/sec. The training time, however, prolongs considerably to accommodate the doubled action dimension of DQN. The number of episodes for convergence increases by 15%.

We do not consider oversaturated traffic conditions in our present experiments. A previous study pointed out that an RL model could not accommodate the traffic demand that exceeds the physical capacity (*8*). The present study tests LOS E that represents the case where the traffic demand approaches to capacity. The test results show that the proposed RL model works well with this near-capacity traffic condition, even though altering turning traffic volumes may make LOS fall to “F” temporarily.

## CONCLUSIONS

This study presents a robust methodology to jointly optimize traffic signals and lane markings. An RL model that belongs to DQN is set up, and a real isolated intersection is chosen as testbed to do so. The state is composed of the queue length and average delay for each lane. The joint action set is designed to incorporate both control and design variables. A commercial traffic simulator, VISSIM v5.0, is selected, and the proposed RL model is trained in the virtual environment that the simulator provides. As a result, the RL model converges in practical computing time even without human filtering of irrational action changes. Three other reference models are employed for comparison. The proposed model outperforms all the reference models in terms of the average number of vehicles passing through the intersection during unit time. In particular, the proposed RL model outperforms RLSIG that optimizes only traffic signals, which clearly confirms the utility of the incorporation of lane marking design.

Despite the promising results from the experiments, the proposed RL model has limitations. The present experiments are conducted only on a single isolated intersection. The present model can be extended to cover a road network that includes multiple intersections. If a MARL model is devised, the present model can be embedded as it is, once a cooperative reward is properly set. Another handicap is that the proposed RL model needs a more formal measure to ensure safety. A more rigorous shield should be integrated with the model to avoid dangerous action changes at inference time. We are delving into resolving this problem and expecting that a robust method could be devised based on the shield RL scheme in the future.

**ACKNOWLEDGMENTS**

This research was supported by the Institute of Information & Communication Technology Planning & Evaluation (IITP) grant funded by the Korea government (MSIT) (No. RS-2024-00361319), and in part by the National Research Foundation of Korea (NRF) Grant funded by the Korean Government (RS-2024-00337956), and in part by the Korea Agency for Infrastructure Technology Advancement (KAIA) grant funded by the Ministry of Land, Infrastructure and Transport (RS-2022-00141102).

**AUTHOR CONTRIBUTIONS**

The authors confirm contribution to the paper as follows: study conception and design: K. Sohn; analysis and interpretation of results: J. Kang and J. Kim; data curation: H. Jun; draft manuscript preparation: K. Sohn. All authors reviewed the results and approved the final version of the manuscript.

**REFERENCES**

1. Shi Y, Guler SI, Zhao J, Zhu J, Yang X. Increasing Signalized Intersection Capacity with Flexible Lane Design. *Transportation Research Part C: Emerging Technologies*, 2025. 173:105054.

2. Zhao J, Liu Y, Wang T. Increasing Signalized Intersection Capacity with Unconventional Use of Special Width Approach Lanes. *Computer-Aided Civil and Infrastructure Engineering*, 2016. 31(10):794-810.

3. Zhao J, Kigen KK, Wang M. Modelling the Operation of Vehicles at Signalised Intersections with Special Width Approach Lane Based on Field Data. *IET Intelligent Transport Systems*, 2020. 14(12):1565-72.

4. Lee J, Chung J, Sohn K. Reinforcement Learning for Joint Control of Traffic Signals in a Transportation Network. *IEEE Transactions on Vehicular Technology*, 2019. 69(2):1375-87.

5. Kim G, Kang J, Sohn K. A Meta–Reinforcement Learning Algorithm for Traffic Signal Control to Automatically Switch Different Reward Functions According to the Saturation Level of Traffic Flows. *Computer-Aided Civil and Infrastructure Engineering*, 2023. 38(6):779-98.

6. Jeon H, Lee J, Sohn K. Artificial Intelligence for Traffic Signal Control Based Solely on Video Images. *Journal of intelligent transportation systems*, 2018. 22(5):433-45.

7. Kim G, Sohn K. Area-Wide Traffic Signal Control Based on a Deep Graph Q-Network (DGQN) Trained in an Asynchronous Manner. *Applied Soft Computing,* 2022. 119:108497.

8. Kang J, Choi Y, Sohn K. Simultaneous Optimization of Traffic Signal Control and Vehicle Platooning Scheme Based on Connected and Automated Vehicle (CAV) Technology. *IEEE Access*, 2025. 13: 99744-99757

9. Shin J, Roh S, Sohn K. Image-Based Learning to Measure the Stopped Delay in an Approach of a Signalized Intersection. *IEEE Access*, 2019. 7:169888-169898.

10. Eom M, Kim BI. The Traffic Signal Control Problem for Intersections: a Review. *European transport research review*, 2020. 12(1):50.

11. Allsop RE. Estimating the Traffic Capacity of a Signalized Road Junction. *Transportation Research,* 1972. 6(3).

12. Little JD, Kelson MD, Gartner NH. MAXBAND: A Versatile Program for Setting Signals on Arteries and Triangular Networks, 1981.

13. Lo HK. A Novel Traffic Signal Control Formulation. *Transportation Research Part A: Policy and Practice*, 1999. 33(6):433-48.

14. Aboudolas K, Papageorgiou M, Kosmatopoulos E. Store-and-Forward Based Methods for the Signal Control Problem in Large-scale Congested Urban Road Networks. *Transportation Research Part C: Emerging Technologies*, 2009. 17(2):163-74.

15. Gartner NH. Development of Demand-Responsive Strategies for Urban Traffic Control. *InSystem Modelling and Optimization: Proceedings of the 11th IFIP Conference Copenhagen*, Denmark, July 25–29, 1983 2005 Oct 1 (pp. 166-174). Berlin, Heidelberg: Springer Berlin Heidelberg.

16. Sun D, Benekohal RF, Waller ST. Multiobjective Traffic Signal Timing Optimization Using Non-dominated Sorting Genetic Algorithm. *InIEEE IV2003 intelligent vehicles symposium. Proceedings (Cat. No. 03TH8683)* 2003 Jun 9 (pp. 198-203). IEEE.

17. El-Tantawy S, Abdulhai B, Abdelgawad H. Multiagent Reinforcement Learning for Integrated Network of Adaptive Traffic Signal Controllers (MARLIN-ATSC): Methodology and Large-scale Application on Downtown Toronto. *IEEE transactions on Intelligent transportation systems*, 2013. 14(3):1140-50.

18. Wiering MA. Multi-Agent Reinforcement Learning for Traffic Light Control. InMachine Learning: *Proceedings of the Seventeenth International Conference (ICML'2000),* 2000. 1151-1158.

19. Arel I, Liu C, Urbanik T, Kohls AG. Reinforcement Learning-based Multi-Agent System for Network Traffic Signal Control. *IET Intelligent Transport Systems*, 2010. 4(2):128-35.

20. Li M, Yin Y, Zhang WB, Zhou K, Nakamura H. Modeling and Implementation of Adaptive Transit Signal Priority on Actuated Control Systems. *Computer-Aided Civil and Infrastructure Engineering,* 2011. 26(4):270-84.

21. Qin X, Khan AM. Control Strategies of Traffic Signal Timing Transition for Emergency Vehicle Preemption. *Transportation research part C: emerging technologies,* 2012. 25:1-7.

22. Noori H, Fu L, Shiravi S. A Connected Vehicle Based Traffic Signal Control Strategy for Emergency Vehicle Preemption. *In Transportation research board 95th annual meeting* 2016. (Vol. 1).

23. Papageorgiou M, Diakaki C, Dinopoulou V, Kotsialos A, Wang Y. Review of Road Traffic Control Strategies. *Proceedings of the IEEE*, 2003. 91(12):2043-67.

24. Wang S, Zhao Y, Zhang S, Wang D, Wang C, Gong B. A Multi-Objective Optimization Method for Single Intersection Signals Considering Low Emissions. *Sustainability,* 2024. 16(9):3522.

25. Stevanovic A, Stevanovic J, So J, Ostojic M. Multi-Criteria Optimization of Traffic signals: Mobility, Safety, and Environment. *Transportation Research Part C: Emerging Technologies*, 2015. 55:46-68.

26. Lam WH, Poon AC, Mung GK. Integrated Model for Lane-Use and Signal-Phase Designs. *Journal of transportation engineering*, 1997. 123(2):114-22.

27. Wong CK, Heydecker BG. Optimal Allocation of Turns to Lanes at an Isolated Signal-Controlled Junction. *Transportation Research Part B: Methodological*, 2011. 45(4):667-81.

28. Xie S, Jiang H. Increasing the Capacity of Signalized Intersections by Allocating Exit Lanes to Turning Movements. *Journal of Advanced Transportation*, 2016. 50(8):2239-65.

29. Li X, Sun JQ. Intersection Multi-Objective Optimization on Signal Setting and Lane Assignment. *Physica A: Statistical Mechanics and its Applications,* 2019. 525:1233-46.

30. Yu C, Ma W, Lo HK, Yang X. Robust Optimal Lane Allocation for Isolated Intersections. *Computer-Aided Civil and Infrastructure Engineering*, 2017. 32(1):72-86.

31. Currie G, Sarvi M, Young B. A new Approach to Evaluating on-Road Public Transport Priority Projects: Balancing the Demand for Limited Road-Space. *Transportation,* 2007. 34(4):413-28.

32. Eboli L, Mazzulla G, Pungillo G. Combining Speed and Acceleration to Define Car Users' Safe or Unsafe Driving Behaviour. *Transportation research part C: emerging technologies,* 2016. 68:113-25.

33. Zhong L, Rong J, Sun XD. Discussion of Lane Width of Elevated Expressway. *Journal of Highway and Transportation Research and Development*, 2006. 23(10):117-9.

34. Zhuang JF, Li ZJ, Ding R, Xiong WL, Zhang HS, Xu J. Vehicles Trajectory Oscillation Characteristics and Passenger Cars' Lane Width for Freeways. *Journal of Transportation Systems Engineering and Information Technology*, 2023. 23(1):324-36.

35. Jiang X, Shang Q. A Dynamic CAV-Dedicated Lane Allocation Method with the Joint Optimization of Signal Timing Parameters and Smooth Trajectory in a Mixed Traffic Environment. *IEEE Transactions on Intelligent Transportation Systems*, 2022. 24(6):6436-49.

36. Papageorgiou M, Mountakis KS, Karafyllis I, Papamichail I, Wang Y. Lane-Free Artificial-Fluid Concept for Vehicular Traffic. *Proceedings of the IEEE,* 2021. 109(2):114-21.

37. Sharma M, Biswas S. Estimation of Passenger Car Unit on Urban Roads: A Literature Review. *International journal of transportation science and technology*, 2021. 10(3):283-98.

38. Abdoos M, Mozayani N, Bazzan AL. Traffic Light Control in Non-Stationary Environments Based on Multi Agent Q-learning. *In2011 14th International IEEE conference on intelligent transportation systems (ITSC)*, 2011. (pp. 1580-1585). IEEE.

39. Abdoos M, Mozayani N, Bazzan AL. Hierarchical Control of Traffic Signals Using Q-learning with Tile Coding. *Applied intelligence*, 2014. 40(2):201-13.

40. Casas N. Deep Deterministic Policy Gradient for Urban Traffic Light Control. *arXiv:1703.09035,* 2017.

41. Bakker B, Whiteson S, Kester L, Groen FC. Traffic Light Control by Multiagent Reinforcement Learning Systems. *InInteractive Collaborative Information Systems*, 2010. (pp. 475-510). Berlin, Heidelberg: Springer Berlin Heidelberg.

42. Mnih V, Kavukcuoglu K, Silver D, Rusu AA, Veness J, Bellemare MG, Graves A, Riedmiller M, Fidjeland AK, Ostrovski G, Petersen S. Human-Level Control Through Deep Reinforcement Learning. *Nature*, 2015. 518(7540):529-33.

43. Dai J, Pan X, Sun R, Ji J, Xu X, Liu M, Wang Y, Yang Y. Safe RLHF: Safe Reinforcement Learning from Human Feedback. *arXiv:2310.12773*, 2023.

44. Bai Y, Jones A, Ndousse K, Askell A, Chen A, DasSarma N, Drain D, Fort S, Ganguli D, Henighan T, Joseph N. Training a Helpful and Harmless Assistant with Reinforcement Learning from Human Feedback. *arXiv:2204.05862*, 2022.

45. Christiano PF, Leike J, Brown T, Martic M, Legg S, Amodei D. Deep Reinforcement Learning from Human Preferences. *Advances in neural information processing systems*, 2017. 30.